\documentclass[sigconf]{acmart}

\acmConference[ICSE 2022]{The 44th International Conference on Software Engineering}{May 21–29, 2022}{Pittsburgh, PA, USA}

\usepackage{multirow}
\usepackage{graphicx}
\usepackage{subfig}
\usepackage{tikz}
\usepackage{threeparttable}

\newcommand{\tool}[1]{{\textit{#1}}}
\newcommand{\param}[1]{{\texttt{{#1}}}}

\usepackage{xspace}
\newcommand{\system}{\textsc{LGDFuzzer}\xspace}
\newcommand{\rvfuzzing}{\textsc{rvfuzzer}\xspace}
\newcommand{\bugs}{range specification bug\xspace}

\newcommand*{\circled}[1]{\lower.7ex\hbox{\tikz\draw (0pt, 0pt)circle (.5em) node {\makebox[0.5em][c]{\small #1}};}}

\begin{document}

\title{Control Parameters Considered Harmful: Detecting Range Specification Bugs in Drone Configuration Modules via Learning-Guided Search}

\author{Ruidong Han}
\email{hanruidong@stu.xidian.edu.cn}
\affiliation{%
  \institution{State Key Lab. for Integrated Service Networks, Xidian University}
  \city{Xian}
  \country{China}
}

\author{Chao Yang}
\email{chaoyang@xidian.edu.cn}
\affiliation{%
  \institution{Xidian University}
  \city{Xian}
  \country{China}
}

\author{Siqi Ma}
\email{siqimslivia@gmail.com}
\affiliation{%
  \institution{The University of New South Wales Canberra}
  \city{Sydney}
  \country{Australia}
}

\author{JiangFeng Ma}
\email{jfma@mail.xidian.edu.cn}
\affiliation{%
  \institution{Xidian University}
  \city{Xian}
  \country{China}
}

\author{Cong Sun}
\email{suncong@xidian.edu.cn}
\affiliation{%
  \institution{Xidian University}
  \city{Xian}
  \country{China}
}

\author{Juanru Li}
\email{mail@lijuanru.com}
\affiliation{%
  \institution{Shanghai Jiao Tong University}
  \city{Shanghai}
  \country{China}
}

\author{Elisa Bertino}
\email{bertino@purdue.edu}
\affiliation{%
  \institution{Purdue University}
  \city{West Lafayette}
  \country{USA}
}

\begin{abstract}
In order to support a variety of missions and deal with different flight environments, drone control programs typically provide configurable control parameters. 
However, such a flexibility introduces vulnerabilities. 
One such vulnerability, referred to as range specification bugs, has been recently identified.
The vulnerability originates from the fact that even though each individual parameter receives a value in the recommended value range, certain combinations of parameter values may affect the drone physical stability.
In this paper
    we develop a novel learning-guided search system to find such combinations, that we refer to as incorrect configurations.
Our system applies metaheuristic search algorithms mutating configurations to detect the configuration parameters that have values driving the drone to unstable physical states. 
To guide the mutations, our system leverages a machine learning predictor as the fitness evaluator.
Finally, 
    by utilizing multi-objective optimization,
    our system returns the feasible ranges based on the mutation search results.
Because in our system the mutations are guided by a predictor, evaluating the parameter configurations does not require realistic/simulation executions. Therefore, our system supports a comprehensive and yet efficient detection of incorrect configurations. 
We have carried out an experimental evaluation of our system. The evaluation results show that the system successfully reports potentially incorrect configurations, of which over $85\%$ lead to actual unstable physical states.
\end{abstract}


\maketitle

\section{Introduction}
\label{sec:intro}


Drones -- flying mini robots, are rapidly growing in popularity. 
They have become essential in supporting the central functions of various business sectors (e.g., motion picture filmmaking) and governmental organizations (e.g., surveillance). 
Because drones are pilotless and have small physical shape and fast speed, they are often used in missions, such as delivery and surveillance, targeting locations difficult or expensive to reach with other means.
However, as drones will be increasingly used also for critical missions, it is important that they be reliable and adaptable to ensure that missions successfully complete.

To achieve reliability and adaptability,  enhanced flight control systems have been developed. Such systems provide large numbers of control parameters that can be configured to modify the flight states of drones, such as linear and angular positions. 
Through parameter adjustment, different configurations can be set and sent to the flight control program. 
Based on such a configuration, the flight control program controls the flight states of the drone to complete the flight mission.
However, the possibility of adjusting parameters introduces certain vulnerabilities (referred to as range specification bugs) arising from the lack of adequate checking of the control parameter values.
Specifically, when setting some particular configurations by selecting parameter values within the ranges provided by the manufacturer, unstable flight states might be triggered, such as trajectory deviation or even drone crash. 

Existing vulnerability detection techniques cannot detect such range specification bugs. 
If the source of the control program is available, static program analysis can be used to detect data/control dependencies~\cite{ma2021orchestration, ma2021fine} and find such bugs. 
However, such an approach works well only for small code snippets. 
If used on large and complex programs,  static analysis would have scalability issues.
To address scalability, taint analysis can be used by tracking input data flow~\cite{cheng2018dtaint,  she2020neutaint, chibotaru2019scalable}, which depends more on input construction.
However, when dealing with very large numbers of control parameters, 
    each with a wide range of values,  analyzing configurations by tainting all the parameter values is time consuming. 
A recently proposed tool, \rvfuzzing~\cite{rvfuzzer}, tries to address such an issue by generating
configuration inputs through fuzzing.
Even though \rvfuzzing is able to reduce the total amount of configurations to be tested, 
    it is still inefficient and unable to provide high-coverage analyses of configurations. 
As a result, it misses incorrect configurations.

Major challenges for detecting range specification bugs include how to validate configurations effectively and how to efficiently search for correct parameter ranges. 
In this work, we address these challenges by developing a learning-guided fuzzing approach specifically designed to detect range specification bugs. 
At a high level, our detection tool, \system, relies on a genetic algorithm (GA)~\cite{ga} and a flight state predictor to detect the configurations that are potentially incorrect. 
Specifically, \system is equipped with three core components, \emph{State Change Predictor}, \emph{Learning-Guided Mutation Searcher}, and \emph{Range Guideline Estimator}. First, we manually collected a list of flight logs, each of which contains a flight state, sensor data, configurations, and a timestamp. 
By using such logs,
    \system trains a state predictor, which is utilized to estimate the state, referred to as reference state, that will be reached by the drone at the next timestamp. 
Simultaneously, \system runs the learning-guided mutation searcher to generate configurations by leveraging the GA. 
Unlike the traditional configuration validation schemes that test the configuration on either a flight simulator or a realistic drone, \system leverages the state predictor to estimate the reference state and infers whether a configuration is correct based on the deviation of the reference state with respect to the expected state.
Finally, \system validates the configurations that are predicted as ``incorrect'' and further generates a valid range for each parameter.

We used \system to analyze the most popular flight control program, \tool{ArduPilot}~\cite{ardupilot}. 
In total, \system validated $46,500$ configurations and labeled $2,319$ as ``incorrect'', out of which $2,036$ incorrect configurations were confirmed. 
Apart from the identified range specification bugs, 
    we also found $564$ zero-day vulnerabilities, referred to as \emph{Incorrect Configuration Tackling bugs}.
These refer to configurations that are detected as incorrect before the drone takes off and,  as a result, the flight is aborted.   
However, the control program will accept these incorrect configurations if they are sent to the control program after the drone has taken off.   
Our analysis also shows that a significant number of incorrect configurations are set in order to enhance adaptability. 
To assist developers and users in building secure flight control programs, \system optimizes the parameter ranges based on a manually set adaptability level. If developers and users prefer higher adaptability, larger parameter ranges will be provided by \system; however the possibility of causing unstable flight would be higher. Otherwise, a smaller range will be set and the flight states of the drone will be more stable.

\vspace{0.1cm}
\noindent
\textbf{Contributions.}

\vspace{0.15cm}
\noindent
(1) We have designed and implemented \system to detect incorrect configurations effectively and efficiently. 
Our system uses a GA to select the ``highly possible'' incorrect configurations and validates configuration correctness by using a deep learning based state predictor. 

\vspace{0.15cm}
\noindent
(2) According to the requirements of reliability and adaptability by developers and users, we designed and implemented a range optimization component in \system to provide the most appropriate parameter value ranges to minimize the possibility of introducing incorrect configurations. 

\vspace{0.15cm}
\noindent
(3) We applied \system to a real-world flight control program and identified $2,036$ incorrect configurations causing unstable flight states. We also verified $106$ incorrect configurations on real-world drones and confirmed that these incorrect configurations cause trajectory deviations or drone crashes.

\vspace{0.15cm}
\noindent
(4) We found a new type of bugs, Incorrect Configuration Tackling bugs, and verified that these bugs also cause unstable flight states. 

\vspace{0.15cm}
\noindent
(5) We have open sourced our \system at \url{https://github.com/BlackJocker1995/uavga/tree/main}; the site makes available the tool, dataset, and video recordings of our tests of incorrect configurations.

\section{Motivation and Challenges}
\label{sec:back}
In this section, 
    we first introduce some information about the flight control programs utilized by drones and the range specification bugs.
We then discuss the challenges in the design of an approach to validate parameter configurations, followed by our solutions to address these challenges.

\subsection{Flight Control Program}\label{back:cp}
During a flight, the ground control station (GCS) communicates with the drone by sending a series of commands to the flight control program.
Before the drone takes off, users can configure the flight control program by adjusting the parameters to manipulate the flight states of the drone (e.g., linear position, angular position, angular speed, velocity, and acceleration).
To ensure that a drone completes its flight mission successfully, the flight control program periodically observes the current positioned flight state and sensor data (e.g., from GPS, gyroscopes, and accelerometers) to estimate a reference state indicating the next state of the drone.
Then the control program generates actuator signals (e.g., motor commands) to move the drone to the reference state.
The positioned state and the reference state need to be close enough, i.e., within a standard deviation. If this is not the case, the drone flight may become unstable, leading to trajectory deviations and crashes.


Although the value ranges for control parameters are typically hardcoded in the control programs and one would expect that all possible combinations of these values be correct, some of the combinations are actually incorrect.
Any configurations triggering unstable flight states are regarded as incorrect. 
The corresponding control parameter ranges are \emph{range specification bugs}~\cite{rvfuzzer}.
Since hundreds of control parameters can be specified by the flight control program, identifying range specification bugs by validating all parameter values is time consuming because some parameters may not affect flight stability. 
Therefore, we focus on the parameters that might affect angular position and angular speed state, because they directly impact the flight attitude and their incorrect values are more likely to cause unstable states.
    
By analyzing the physical impact on drones, we define five unstable flight states:

\vspace{0.15cm}
\noindent    
\textbf{Flight Freeze.}
A drone is required to keep moving unless the flight control program generates a signal to keep the drone stationary or to instruct the drone to land.
However, incorrect configurations may lead the drone to accidentally freeze at a waypoint, when moving forward/backward, or to wander around a position within a minimum range. 
To determine whether a flight is frozen,
    we calculate the movement distance between the previous and the current positions within a time interval. 
If the distance is less than a threshold, the flight state is regarded a frozen, i.e.,``flight freeze''.

\vspace{0.15cm}
\noindent
\textbf{Deviation.}
According to the actuator signals generated by the flight control program, the drone will be driven to achieve the reference state as close as possible.
In practice, however, the positioned state may not always match the reference state, that is, the drone is deviating from the expected trajectory.
When the deviation is small, the control system can generate an actuator signal based on the differences between the current positioned state and the reference state.
However, an incorrect configuration may trigger a significant deviation. 
Such a deviation may lead to an erroneous trajectory from which the drone cannot return back to the correct trajectory.
If the deviation exceeds a given threshold, we consider the flight state as unstable, i.e., a ``deviation'' state.


\vspace{0.15cm}
\noindent
\textbf{Drone Crash.}
For general deviations, the drone can still land safely even though not at the expected location. 
A worse case is a deviation leading the drone against an object and eventually crash. 

\vspace{0.15cm}
\noindent
\textbf{Potential Thrust Loss.}
By driving the drone motor, the flight control program uses motor power to adjust the drone to the reference state.
Nonetheless, the adjustment that can be done by the drone motor is limited.
If an incorrect configuration is set, the drone may not be able to move close to the reference state even when saturating the motor up to 100\% throttle.
Remaining in such an incorrect state can cause a decrease in the drone flight altitude and attitude, or even a crash. 

\vspace{0.15cm}
\noindent
\textbf{Incorrect Configuration Tackling.}
Before taking off, the flight control program validates the configuration and determines whether it will trigger unstable flight states. (i.e., the configuration is incorrect).
When one or more incorrect parameters of the configurations are identified, the control program displays a warning message and aborts the taking-off operation.
However, if these configurations, flagged as incorrect, are set after taking off, they can still be accepted by the flight control program.
As a consequence, the drone may end up in some unstable states.

\subsection{Challenges}\label{sub:challenges}
In order to validate all configurations and detect range specification bugs, 
    the following challenges must be addressed:

\vspace{0.15cm}
\noindent
\textbf{Challenge \uppercase\expandafter{\romannumeral1}: How to validate configurations effectively?}
Approaches proposed for analyzing flight control programs are generally based on static program analysis techniques that explore control and data dependencies~\cite{MAYDAY,Believing,zhu2019new}.
However, such techniques are not suitable for validating configuration because large numbers of specified parameter values need to be analyzed.
Unlike conventional bug detection techniques that statically analyze only small code snippets, the entire flight control program needs to be analyzed in order to achieve high code coverage. 
The reason is that different parameter values often result in execution flows involving very different portions of the control program. 
Unfortunately, flight control programs have huge sizes (e.g., over 700K lines of code) and complex control and data dependencies,
Therefore, it is critical that the approach designed for configuration validation be effective, that is,  able to provide high coverage of all possible parameter values.

\vspace{0.15cm}
\noindent
\textbf{Challenge \uppercase\expandafter{\romannumeral2}: How to conduct an efficient configuration validation?}
Referring to the unstable flight states defined in Section~\ref{sec:back}, we need to validate each configuration through either a realistic or simulated flight execution. 
Because of the large number of control parameters, each with its value range, changing the parameter values to generate configurations and validating all these configurations is inefficient
Completing the entire validation procedure may then end up requiring hours. 
Therefore, existing approaches~\cite{tartler,halin2019test}, which analyze all possible configurations, are not suitable.
    
An alternative approach is to use fuzzing~\cite{rvfuzzer} combined with a binary search~\cite{knuth1971optimum} to reduce the search space of the combinations to be analyzed.
However, although the search scope is narrowed, the execution time increases because each fuzzing iteration needs to wait until the validation feedback of the previous configuration is obtained.

\vspace{0.15cm}
\noindent
\textbf{Challenge \uppercase\expandafter{\romannumeral3}: How to balance the requirements of drone adaptability and flight stability?}
The flight control program supports different configurations to adapt to different flight missions. 
When high adaptability is required, each control parameter must have a large value range to adapt to different scenarios.
As a result, the number of incorrect configurations will be higher, which will then affect flight stability.
On the other hand, when parameters have small value ranges, the possible configurations are limited, which reduces adaptability. Hence, complex flight missions cannot be carried out.
Identifying proper value ranges is thus challenging.

\subsection{Solutions}
\label{sub:solution}
We now introduce our approaches to the above challenges.

\vspace{0.15cm}
\noindent
\textbf{Solution I: Grey-box based fuzzing.}
Since dependencies of the flight control program are complex, static program analysis techniques are ineffective.
Our approach is instead to conduct grey-box based fuzzing by setting various configurations and validating the corresponding flight states.
In particular, we apply a GA to carry out fuzzing. 
The algorithm first selects some parameter values and validate the resulting parameter configuration by analyzing the impact of the configuration on the flight states.
After that, referencing the validation result, the algorithm conducts a mutation to select more incorrect parameter values and set new configurations to validate,
We repeat this process to search more incorrect configurations.
This approach address challenges I and II.


\vspace{0.15cm}
\noindent
\textbf{Solution II: Flight state prediction.}
Although the GA and the fuzzing reduce the search range of parameter values, the number of combinations is still huge.
Therefore, validating all the corresponding configurations through realistic/simulation execution is highly inefficient. 
In response, we designed a state generation approach that leverages a machine learning algorithm to train a state predictor.
Instead of validating configurations through a realistic/simulation execution, the state predictor takes each configuration as input and predicts the potential flight state to guide the mutation. 
Such an approach requires much less time than a realistic/simulation execution. 
It is important to note that data labeling and predictor generation are one-time costs. 
Hence it is still more efficient to conduct a prediction rather than a realistic/simulation execution. 
The reason is that configuration validation has to be executed iteratively when each new configuration is generated by the mutation algorithm.
This solution addresses Challenge II.


\vspace{0.15cm}
\noindent
\textbf{Solution III: Multi-objective optimization.}
To balance adaptability and stability, we utilize a multi-objective optimization approach. 
Our approach estimates multiple feasible range guidelines according to the detected incorrect configurations. 
The optimization target is to eliminate the incorrect configurations (improve stability) while providing a wider range for parameters (improve adaptability).
Each solution of the multi-objective optimization (i.e., range guideline) is the best solution in specific conditions, i.e., the optimal balance of adaptability and stability. Our approach allows users to choose an appropriate range from multiple range guidelines based on their requirements.
This approach addresses Challenge III. 



\section{\system}
\label{sec:design}
In this section, we present the design of \system, our system for finding range specification bugs drone control programs through learning-guided mutation search.
We first present an overview of its architecture and then the detailed design of its three components.

\subsection{Overview of \system}
\begin{figure}[ht]
  \centering
    \includegraphics[width=\columnwidth]{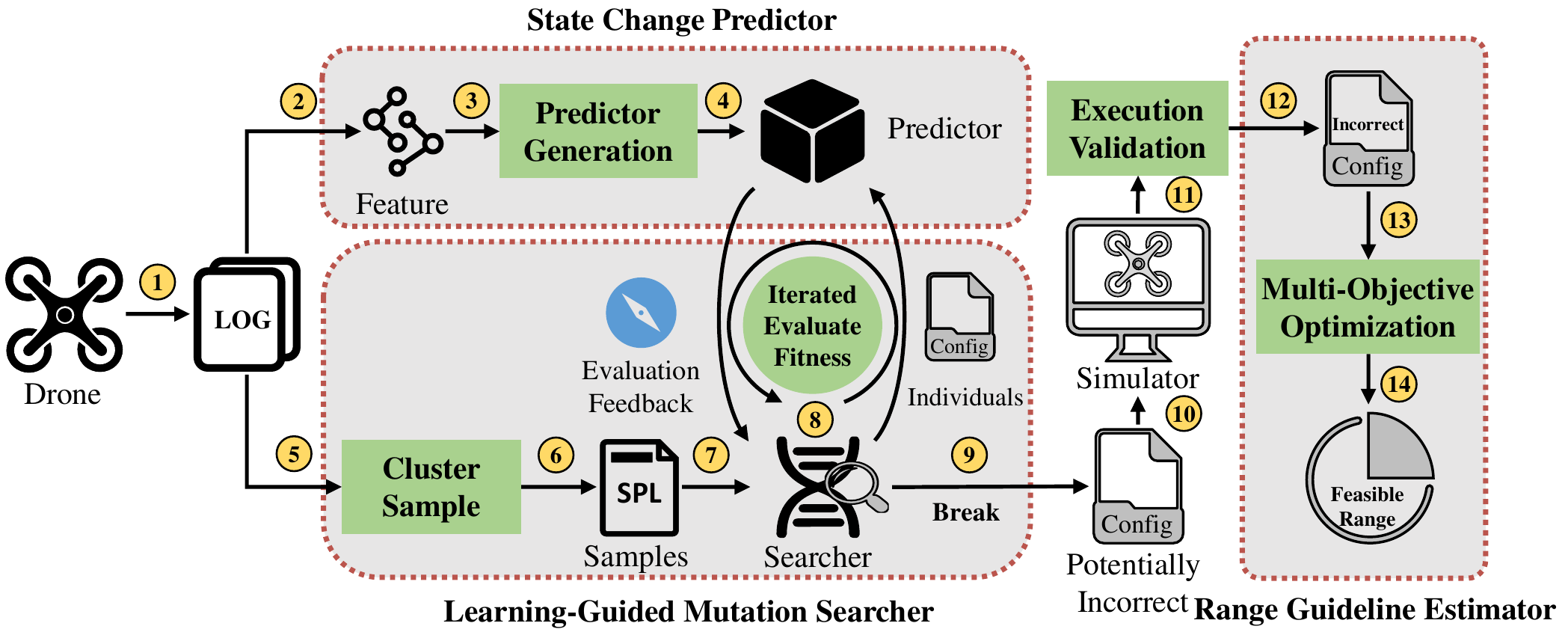}
\caption{Overview of \system.}
\label{fig:overview} 
\end{figure}

\system (see Fig.~\ref{fig:overview}) contains three components: 
    \emph{state change predictor}, a module to predict the flight state and assess whether a configuration will lead to unstable states;
    \emph{learning-guided mutation searcher}, a GA search module to detect incorrect configurations;
    \emph{range guideline estimator}, a module to provide secure parameter range guidelines.

\system relies on log files that we generate by repeatedly executing drone flight missions (\circled{1}).
The log data is split into two parts, one is used by the \emph{state change predictor} to generate features (\circled{2}), and the other by the \emph{learning-guided mutation searcher} to carry out cluster sampling (\circled{5}).
After that, the features are used to generate a predictor (\circled{3}\circled{4}).
The searcher clusters data and selects representatives samples from each cluster (\circled{6}).
The searcher then uses each representative sample to find out the corresponding incorrect configurations (\circled{7}).
It iteratively mutates the configuration and uses the predictor to evaluate which configuration is more likely to cause unstable states (\circled{8}).
When the iteration stop condition is satisfied,  the searcher merges the search result of each sample and generates a potentially incorrect configuration set (\circled{9}).
Then, these potentially incorrect configurations are validated  by a simulation (\circled{10}\circled{11}\circled{12}).
Finally, by using the validation results, the estimator uses a multi-objective optimization to generate multiple feasible range guidelines that balance availability and stability under specific conditions (\circled{13}\circled{14}).
    

\subsection{Flight Log Generation}
Since there is no standard data set for testing drones,
    in order to extract features for the predictor and generate data for the searcher, 
    we manually fly a drone through a simulator to generate a number of flight logs. 
A flight log contains multiple entries and each entry consists of state information, sensor data, configuration, and a timestamp index. 
Each flight is set up with the same flight test mission, \tool{AVC2013}~\cite{avc}, which is often used to test the drone mission execution capabilities. 
Such a flight mission is repeatedly executed  with different configurations.
We record all flight logs but discard those causing unstable states.
Because unstable state data is uncontrollable and complex (compared to stable state data), 
    dropping them prevents them from affecting the training of the predictor. 
In our experiments, we recorded
    $308,533$ system log entries.

\subsection{State Change Predictor}
\label{sub:design-learning}
As mentioned in Sec.~\ref{sec:back}, the control algorithm estimates the next reference state according to the current positioned state and sensor data.
Based on this input-output control process, we leverage a machine learning (ML) predictor to emulate this input-output relationship and assess the impact of configurations.
There are several reasons for using such a ML predictor.
First, the diversity and flexibility of ML predictors make them appropriate to emulate the non-linear input-output control process.
Second, a ML predictor can estimate the next reference state while accurately reflecting the deviation caused by incorrect configurations.
Specifically, for flight data, the learning algorithm trains a predictor that takes the positioned state, sensor data, and configuration as input to predicts (estimates) the next reference state.
Like the control algorithm, 
a large deviation between the predicted and actual reference states indicates that the flight state could become unstable.
Two steps are executed to train the predictor: 
\emph{feature extraction} and \emph{predictor generation}.

\vspace{0.15cm}
\noindent
\textbf{Feature Extraction.}
To train a predictor by using our dataset,
    the state, sensor data, and configuration must be extracted from the original logs and represented according to a fixed feature vector format.
Since we focus on the unstable states that influence the angle (i.e., position and speed), 
    our predictor considers: (i) angular position and angle speed of the state; (ii) data obtained from 
    gyroscopes and accelerometers.     
A feature consists of a state unit $a$,
    a sensor data unit $s$,
    and a configuration unit $x$.
For ease of presentation, 
    we refer to the combination of a specific state unit and sensor data unit as contexts $c=\{a,s\}$.    
Finally,
    a feature consists of those three units with a timestamp and is normalized to a vector $v\{c,x\}$.

\vspace{0.15cm}
\noindent
\textbf{Predictor Generation.}
According to previous research~\cite{stock,highway,selvin2017stock} the Long Short-Term Memory (LSTM)~\cite{gers1999learning} technique can efficiently fit complex input/output of non-linear models.
Therefore, we use \tool{LSTM} as predictor for the next reference state. 
Specifically, for a pre-processed feature vector $v_i=\{c_t,x_t\}$ with timestamp $t$, \tool{LSTM} takes a number $h$ of consecutive vectors with timestamp before $t$ as input (i.e., $V\{v_{t-h-1},...,v_{t-1},v_{t}\}$), 
    and returns the maximum conditional probability prediction of the next reference state units $a_{t+1}'$. We show in Sec.~\ref{sec:eval} how we determine the best values for $h$. 
In the training stage, the weight of the predictor is iteratively updated to ensure that the predicted reference state $a_{t+1}'$ is closer to the ground truth state $a_{t+1}$.
We use the \tool{Mean Squared Error (MSE)}~\cite{4775883} loss for training.

\subsection{Learning-Guided mutation searcher}\label{sub:design-mutation}
To search the incorrect configurations,
    we use a GA, that is,
    a metaheuristic search relying on biologically inspired operators such as mutation, crossover, and selection.
Initially,
    as incorrect configurations may only produce unstable states in specific contexts (i.e., state and sensor data),
    our system conducts searches for multiple specific contexts to find incorrect configurations.
We split the  data from logs into multiple segments $C_i = \{c_{i},c_{i+1},...,c_{i+h}\}, i~mod~(h+1) = 0$,
    where $i$ is the number of data. 
But considering that there are similarities among segments, searching for each one will generate a huge number of duplicate results. To address such issue, we cluster the segments and sample from these clusters in order to reduce redundancy while maintaining diversity.
We leverage the \tool{meanshift}~\cite{cheng1995mean},
    a probability density-based non-parametric adaptive clustering algorithm,
    to cluster segments and randomly sample $m$ representative examples from each cluster for the subsequent search.
Then,
    for each sample segment,
    the searcher carries out a GA search to explore incorrect configurations by iterative mutation, crossover, and selection.
When the searcher has collected all the incorrect configurations for each sample,
    it merges and de-duplicates them as a unique set, referred to as {\em set of potentially incorrect configurations}.
    
In what follows we provide details on the \textit{Fitness Evaluation Function}, used to evaluate the fitness of a configuration,  and the \textit{Searching Process}.

\vspace{0.15cm}
\noindent
\textbf{Fitness Evaluation Function.}
The GA search applies fitness to quantify, by using the predictor, how much deviation a given configuration may cause.
Intuitively, the fitness evaluation function returns the probability of deviation for a configuration. 
Specifically,
    assume that the current search is carried out for the context segment $C\{c_1,...c_{h},c_{h+1}\}$;
    the function selects $C'\{c_1,c_2,...c_{h}\}$ to be used as input to the prediction model and $a_{h+1}$ of $c_{h+1}$ to be used as ground truth for calculating the deviation.
When evaluating a configuration $x\{x_1,...,x_D\}$ ($D$ is the number of parameters),
    the function merges it with the segment to create features $V\{\{c_1, x\},...,\{c_h,x\}\}$.
Then, such features are given as input to the predictor, which returns a predicted reference state $a'_{h+1}$.
The fitness is the L1-distance $\Vert a_{h+1} - a'_{h+1} \Vert$ between the predicted state and the ground truth state.
The goal of the search is then to find incorrect configurations, which are predicted to maximize the fitness, that is, the deviation (and thus maximize the probability of causing unstable states). 

\vspace{0.15cm}
\noindent
\textbf{Searching Process.}
For each context segment sample, the searcher first initializes a population whose individuals are configurations,
    and the parameter values of each configuration are set to their default values.
We assume that the population size is $NP$ and the maximum number of iterations is $G_{max}$.
The search process iteratively mutates and updates the population as follow:

In the $g$-th generation iteration ($g\in[1,G_{max}]$),
    the searcher first mutates the current population $pop_g\{x_{1,g}...,x_{NP,g}\}$ and generates a variant population $pop_v\{y_{1,g},...,y_{NP,g}\}$.
Each configurations of the variant population is obtained as follows:
\begin{equation}
    y_{i,g} = x_{i,g} + F * (x_{best,g} - x_{i,g}) + F * (x_{r1,g} - x_{r2,g})
\end{equation}
    where $x_{r1/r2,g}$ are random configurations,
    $x_{best,g}$ is the best fitness configuration, and
    $F$ is the scaling factor.
    
Then,
    the $pop_g$ is crossed over with the $pop_v$ to produce a new experimental population $pop_e\{e_{1,g},...,e_{NP,g}\}$, whose $i$-th configuration is $e_{i,g}\{e_{i1,g},e_{i2,g},...,e_{ij,g}\}$,
    where $j\in[1,D]$.
The parameter values $e_{ij,g}$ of each configuration are calculated as follows:
\begin{equation}
    e_{ij,g}=\begin{cases}
y_{ij,g}, & if~rand(0,1) < CR~or~j=j_{rand}\\
x_{ij,g}, & otherwise
\end{cases}
\end{equation}   
    where $j_{rand}$ is a random integer in $[1,D]$,
    $CR$ is the crossover rate,
    $x_{ij,g}$ is the $j$-th parameter value of $i$-th configuration in $pop_g$, 
    and $y_{ij,g}$ is the $j$-th parameter value of $i$-th configuration in $pop_v$.
    
Then, the searcher evaluates the fitness of each configuration both in $pop_g$ and $pop_e$. 
According to their fitness, the searcher selects some configurations from the population to obtain the next generation of population $pop_{g+1}\{x_{1,g+1},...,x_{NP,g+1}\}$ by using the following selection function:
\begin{equation}
    x_{i,g+1}=\begin{cases} 
e_{i,g}, & if~f(e_{i,g})<f(x_{i,g})\\
x_{i,g}, & otherwise
\end{cases}
\end{equation}   
    where $f$ is the fitness evaluation function.
    
Finally,
    if the fitness of each configuration in the population does not longer increase or the maximum number $G_{max}$ is reached,
    the search stops the mutation.
We select the top $10$ highest-fitness configurations from the final generation population as incorrect configurations.

\subsection{Execution Validation}    
The predictor classifies a configuration as incorrect if the configuration has a high probability of causing unstable states. However the set of all such configurations
need to be validated to confirm that they actually cause unstable states.
We thus simulate the flight and use a monitor to observe which potential incorrect configurations actually lead to unstable states during the simulated execution.
Specifically, the drone is set with potentially incorrect configurations to perform the \tool{AVC2013} flight mission.
For flight freeze, if drone moves at  distances always less than $0.5$ meters in $15$ seconds, the monitor identifies this as a flight freeze.
For deviations, if the flight deviation continues to be $15$ times higher than $1.5$ meters, the flight is considered a deviation.

\subsection{Range Guideline Estimator}
The range guidelines provided to the users should consider stability and adaptability, while at the same time eliminate discrete incorrect configurations and reserving relatively complete space for each parameter.
Meanwhile, as we cannot ensure the stability of unverified configurations, the estimation of range guidelines should reference the obtained validation results and refrains from considering incorrect configurations.
Therefore, we leverage a multivariate optimization to determine the suitable range guideline.
If all parameter values in a configuration are within the range specified by the guideline,  we consider the configuration to be covered.
Our system attempts to estimate the range guideline ($Range'$) based on the following optimization problem:
\begin{equation}
\begin{cases}
        min~f_1 = \frac{num(R_{incorrect } \in Range')}{num(R \in Range')}   \\
       max~f_2 = num(R \in Range')  \\
       s.t. &\\
       Range' \in OriginRange
\end{cases}
\label{eq1}
\end{equation} 
The optimization problem has two objectives: (a) to maximize the number of validated configurations $R$ covered by the guideline; (b) to minimize the coverage of incorrect configurations $R_{incorrect }$.
If we shrink the allowed ranges to avoid incorrect configurations,
    the range of each parameter value would also shrink and vice versa.
Therefore, instead of defining strict ranges for each parameter,
    the system solves this optimization problem with multiple constraints to obtain a diverse group of \tool{Pareto} solutions.
These \tool{Pareto} solutions form a boundary consisting of the best feasible solutions,
    allowing users to select the best configuration according to their requirements.
The system provides this range guideline mainly based on the following considerations:
(1) As the number of parameter increases,
    it would be difficult to completely rule out insecure parameter values as the secure parameter ranges may not be continuous.
(2) While stability is paramount,
    users may be willing to incur some minor risk for better controlling flexibility or availability.

\section{Evaluation}
\label{sec:eval}
We assessed \system by considering its effectiveness and efficiency in validating configurations.
The following research questions (\textbf{RQs}) were answered:

\begin{itemize}
 
\item \textbf{RQ1: Effectiveness.} How many incorrect configurations are detected by \system?

\item \textbf{RQ2: Adaptability.} Can \system provide the most suitable value ranges for parameters with minimum incorrect configurations?

\item \textbf{RQ3: Enhancement.} How do the mutation and the predictor help improve configuration validation?

\end{itemize}

\subsection{Experiment Setup}
\begin{figure}[htb]
\centering{
\quad
\subfloat[CUAV ZD550]{\includegraphics[width=0.25\linewidth]{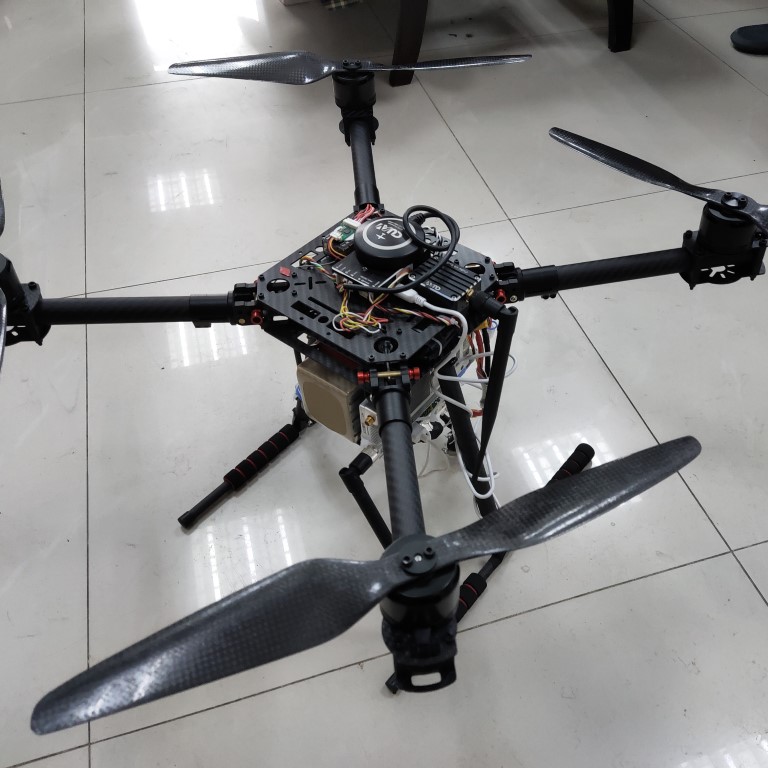}}\quad\quad
\subfloat[AMOVLab Z410]{\includegraphics[width=0.25\linewidth]{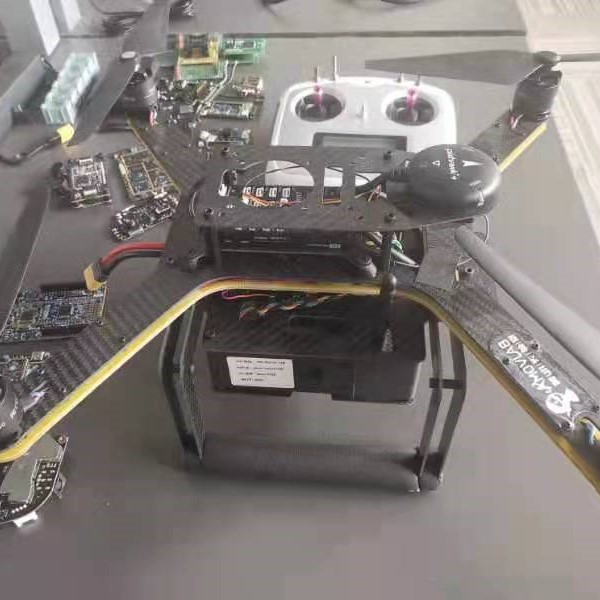}}\quad\quad
\subfloat[Airsim]{\includegraphics[width=0.25\linewidth]{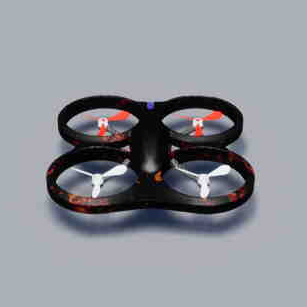}}\quad\quad}
\caption{Real and virtual drone vehicles used for experiments.}
\label{fig:pix}
\end{figure}

We applied \system to an open source flight control program, 
    \tool{ArduPilot}(4.0.3)~\cite{ardupilot}, 
    which is widely used by drone manufacturers such as \tool{Parrot}, \tool{Mamba}, and \tool{Mateksys}~\cite{handware}.
To validate configurations specified in \tool{ArduPilot}, 
    we utilized three experimental vehicles (shown in Fig.~\ref{fig:pix}) for testing, including two drones with \tool{Pixhawk}~\cite{meier2011pixhawk} (i.e., \tool{CUAV ZD550} and \tool{AMOVLab Z410}) and a drone simulator (i.e., \tool{Airsim}~\cite{airsim2017fsr}).
    
According to the control parameter descriptions provided by the manufacturer, 
    we selected $20$ parameters that may affect flight angular position and angular speed.
We provide details about these parameters in Appendix~\ref{app:a}, 
    including parameter name, 
    specified value range, 
    default value, 
    and parameter description.

The predictor and the GA searcher are implemented in Python.
Concerning the GA, its evolutionary stagnation judgement threshold is set to 0.1, 
    the number of representatives $m$ is set to $3$.
    and the maximum number of evolutionary generations is set to $200$. 
We further set the size of initial population to $1,000$ (i.e., $NP=1,000$) and the scaling factor $F$ to 0.4.



\subsection{RQ1: Effectiveness}
\label{ex1}
To evaluate whether \system identifies incorrect configurations accurately, 
    we first assess the prediction accuracy of the predictor in the flight state predictor phase. 
Then we validate whether the predicted incorrect configurations can impact flight stability. 

\vspace{0.2cm}
\noindent
\textbf{State Prediction.}
We labeled 1,500 configurations manually by sending the configurations to \tool{Airsim} and recording the resulting flight states.
Then we use previously collected log data about stable states to train and evaluate predictors with different input lengths $h$.
Specifically, our experiment first extracts the context (i.e., state and sensor data) segments from log data.
Then, each labeled configuration is randomly merged with a segment. We use the same method in \textit{Fitness Evaluation Function} to calculate the deviation for each configuration.
If the deviation is greater than a threshold, the configuration is considered unstable; it is considered stable, otherwise. 
In particular, we use the maximum deviation in the predictor training process, that is, the maximum deviation from the stable flight data, as the threshold.
This experiment tests the accuracy, precision, and recall of predictors.
The results are reported in Table~\ref{tab:cross}.








\begin{table}[ht]
\caption{Predictor accuracy, precision and recall for different input lengths.}
\label{tab:cross}
\centering
\begin{threeparttable}
\begin{tabular}{cccccc}
        \hline
        $h$  & Accuracy & Precision & Recall & \textbf{DC} & \textbf{DI} \\
        \hline

        3  & 0.8650 & 0.8786 & 0.9818 & 9.7643 & 14.6818 \\

        4  & 0.8689 & 0.8778 & 0.9882 & 9.6246 & 14.6508 \\

        5  & 0.8546 & 0.8777 & 0.9695 & 9.8126 & 14.7393\\

        6  & 0.8662 & 0.8769 & 0.9553 & 9.6246 & 14.6508\\

        \hline
\end{tabular}
\begin{tablenotes}
\footnotesize
\item[*] $h$ is the input length of the model, \textbf{DC} is the average deviation of correct configurations, \textbf{DI} is  the average deviation of incorrect configurations.
\end{tablenotes}
\end{threeparttable}
\end{table}

In general, the results show that our predictor is robust for different input lengths.
The deviation values, observed from the experiments, of incorrect configurations are larger than the values of the correct configurations.
That is the reason why we can use the predictor in the GA search to drive configurations to a higher deviation.
In addition, the experiment results show that the recall of the prediction does not linearly increase when $h$ exceeds $4$. 
Therefore,
    we choose the predictor with the best recall i.e., $h=4$ as the input length to carry out the subsequent experiments.

\begin{figure}[htb]
\centering{
\includegraphics[width=0.8\linewidth]{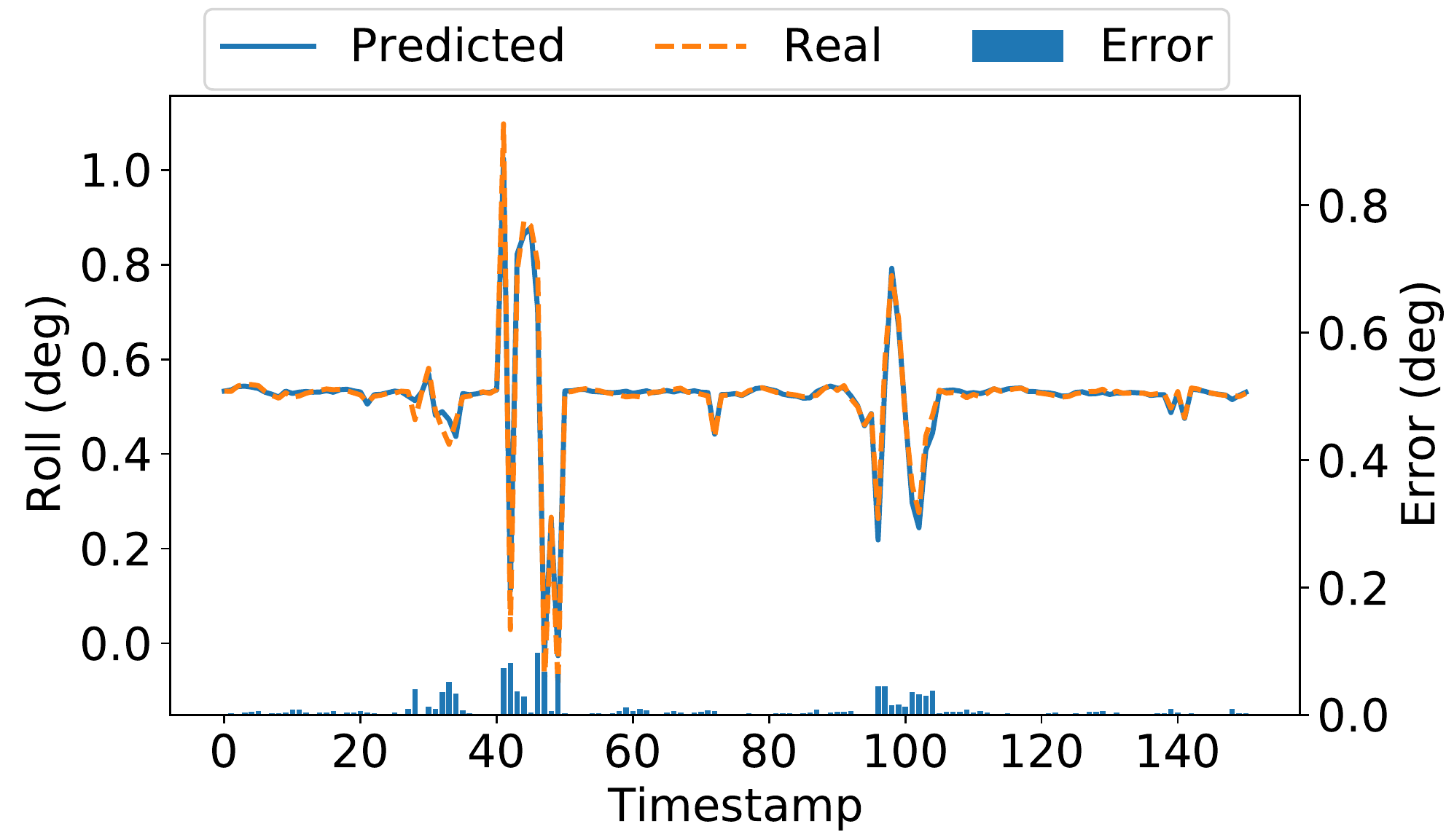}
}
\caption{Match between real behavior and  prediction.} 
\label{fig:diff}
\end{figure}

To assess the prediction accuracy for state changes,
    we send the configurations to \tool{Airsim} and record the corresponding flight data.
Then, we randomly select $150$ consecutive features from the data and give them as input to the predictor to predict the state change and then compare it with the ground truth state.
The results in Fig.~\ref{fig:diff} show that the predictions closely match the real states (an example of angle roll).
In the figure, 
    the dotted curve denotes the actual states, 
    the solid curve denotes the predicted states,
    and the histogram at the bottom (the blue bar) indicates the prediction errors as differences between the two curves.
The small error shows that the trained predictor is able to accurately predict flight state changes.

\vspace{0.2cm}
\noindent
\textbf{Configuration Validation.}
Given the predicted incorrect configurations, 
    we validate them through realistic/simulation execution.
Since the incorrect configurations may cause drone crash, we examine all of them on \tool{Airsim}.
For the $465$ samples obtained by clustering and sampling,
    \system searches out $2,319$ unique incorrect configurations.
Finally,
    after a validation,
    $2,036$ configuration of $2,319$ are marked as truly incorrect resulting in $500$ \emph{Deviations},
    $2$ \emph{Flight Freeze},
    $2$ \emph{Crashes},
    $564$ \emph{Incorrect Configuration Tackling},
    and $968$ \emph{Potential Thrust Losses}.
    

\subsection{RQ2: Adaptability}
\begin{figure}[htb]
   \centering
    \includegraphics[width=0.95\columnwidth]{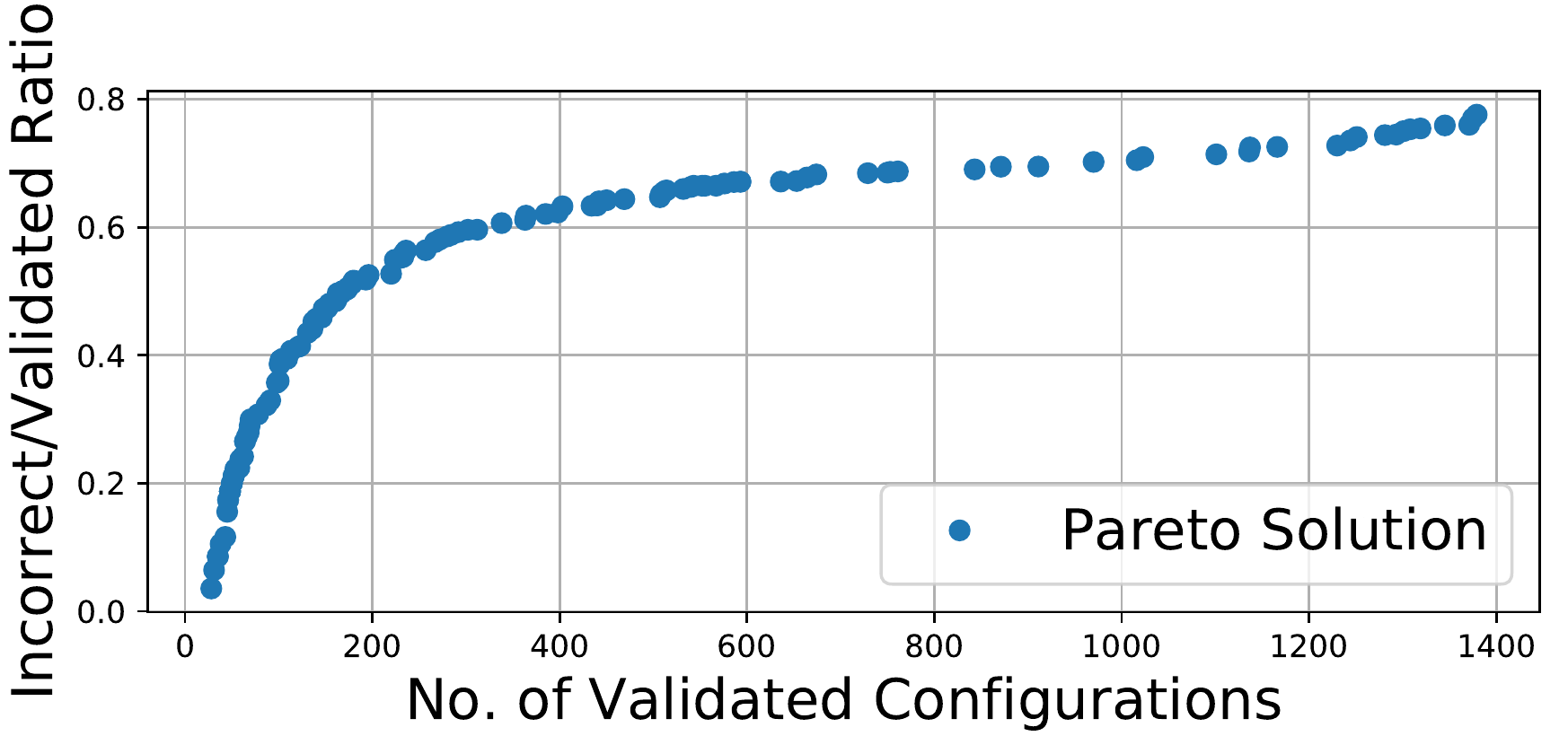}
\caption{\tool{Pareto} frontier solution.} 
\label{fig:pareto} 
\end{figure}

In this experiment, 
    we use the previous validation results about incorrect configurations to estimate the range guidelines,
    and discuss how they balance the stability and adaptability. 
With reference to the validation results,
    the estimator finds out $143$ \tool{Pareto} solution results (see Fig.~\ref{fig:pareto}). In the graph,
    the horizontal axis represents the number of validated configurations covered by the range guideline,
    and the vertical axis represents the ratio of incorrect configuration in the range guideline.  
Each \tool{Pareto} solution represents a feasible solution (i.e., range guideline) satisfying specific constraints (i.e., adaptability and stability constraints).

\begin{table*}[ht]
\caption{Examples of feasible range guideline }
\label{tab:shrink}
\centering
\begin{threeparttable}
\begin{tabular}{c|ccc|ccc|ccc}
        \hline
         \multirow{2}{*}{Parameter}  & \multicolumn{3}{c|}{\textit{Guideline} \romannumeral1} & \multicolumn{3}{c|}{\textit{Guideline} \romannumeral2} & \multicolumn{3}{c}{\textit{Guideline} \romannumeral3} \\
         \cline{2-10}
         ~ & \textbf{L} & \textbf{U} & Reduce& \textbf{L} & \textbf{U} & Reduce& \textbf{L} & \textbf{U} & Reduce \\
         
        \hline
     
         PSC\_POSZ\_P  &  1.00 &  2.57   & -21.5\%&  1.00 & 2.59  & -20.5\%  & 1.00 & 2.97   & -1.5\%\\
        
         ATC\_ANG\_RLL\_P  &  0.92 &  11.4   & -12.6\%&  0.88 & 11.73  & -9.5\%  & 1.02 & 11.69  & -11.0\%\\
        
         ATC\_ANG\_PIT\_P &   1.22 &  11.69   & -12.7\%&  0.94 & 11.78  & -9.6\%  & 1.00 & 11.93  & -8.9\%\\
        
         ATC\_RAT\_PIT\_P &   0.035 &  0.39   & -27.5\%&  0.035 & 0.375  & -30.6\%  & 0.01 & 0.36  & -28.5\%\\
        
         ATC\_RAT\_PIT\_I &   0.13 &  1.92   & -10.0\%&  0.11 & 1.93  & -8.5\%  & 0.10 & 1.96  & -6.5\%\\
    
         ATC\_RAT\_YAW\_I &   0.01 &  0.69   & -31.3\%&  0.01 & 0.72  & -28.2\%  & 0.01 & 0.95 & -5.0\%\\
        
         WPNAV\_SPEED &   250 &  1550   & -34.3\%&  250 & 1850  & -19.1\%  & 100 & 1800  & -14.1\%\\
         
         WPNAV\_SPEED\_UP &   150 &  650   & -49.4\%&  50 & 650  & -39.3\%  & 50 & 650 & -39.3\%\\
        
        \hline
        \textbf{I},\textbf{C},\textbf{V} & 
        \multicolumn{3}{c|}{1,27,28} & 
        \multicolumn{3}{c|}{29,62,91} & 
        \multicolumn{3}{c}{133,103,236} \\

        Coverage(covered/total validation) & 
        \multicolumn{3}{c|}{0.05\%,9.5\%,1.2\%} & 
        \multicolumn{3}{c|}{1.4\%,21.9\%,3.9\%} & 
        \multicolumn{3}{c}{6.5\%,36.3\%,10.1\%} \\
        
        \hline
\end{tabular}
\begin{tablenotes}
\footnotesize
\item[*] \textbf{L} is range lower bound,
\textbf{U} is range upper bound,
reduced is calculated relative to original range,
\textbf{I} is the number of incorrect configurations covered by the range guidance,
\textbf{C} is the number of correct configurations covered by the range guidance,
\textbf{V} is the number of validated configurations covered by the range guidance.
\end{tablenotes}
\end{threeparttable}
\end{table*}

For instance, we select some range guideline examples in Table~\ref{tab:shrink} and further analyze their stability and adaptability.
To illustrate the modified guidelines generated by \system, we select several parameters out of $20$ ones to show their details.  
The table shows three examples in which stability decreases and the configurable space (i.e., adaptability) increases. 
\textit{Guideline} \romannumeral1~avoids the majority of incorrect configurations.
It covers $28$ validated configurations, only one of which causes an unstable state; this means high stability.
However, compared with the original parameter ranges, this guideline reduces too much the available space, which results in low adaptability.
In contrast, \textit{Guideline} \romannumeral3~reserves relatively complete ranges for the parameters.
It covers $236$ validated configurations but more than half of them ($133$) are incorrect,
    which results in low stability.
\textit{Guideline} \romannumeral2~is an intermediate choice, 
    covering $91$ validated configurations where only  $29$ are incorrect. 
    
If users have more stringent stability requirements, they can use the lower error ratio range guideline, at the cost of limiting the configuration space and thus being unable to satisfy other flight requirements.
On the contrary, if users have to satisfy special mission requirements (e.g., the mission is a time-limited task, or it needs a large flight angle to reach the target speed), they may consider sacrificing a bit the stability in order to improve adaptability. 
In fact, 
    they can choose an appropriate range guideline from the \tool{Pareto} solutions according to their stability and adaptability requirements.

\subsection{RQ3: Improvement}
To show the advantages of our system,
    we experimentally compare it with \rvfuzzing~\cite{rvfuzzer}, a recent approach for searching \bugs.
\rvfuzzing relies on the \tool{One-dimensional Mutation} and \tool{Multi-dimensional Mutation} search to detect incorrect configurations.
By using \tool{One-dimensional Mutation},
    centered on the default parameter values,
    \rvfuzzing separately conducts binary searches to narrow the upper and lower bounds until a midpoint value is obtained that does not any longer cause unstable states.
In the \tool{Multi-dimensional Mutation} multiple parameters are considered,
    each of which determines a novel binary mutation search configured with different extreme values (i.e., maximum and minimum) of the other parameters.
All these experiments are based on the six parameters utilized in \rvfuzzing (see Appendix~\ref{app:a}). In the experiments
    we consider the unstable states listed in Section~\ref{back:cp}.

\vspace{0.15cm}
\noindent
\textbf{Comparison with Respect to Missed Incorrect Configurations.}
In the \tool{One-dimensional Mutation}, the optimal solution may not be consistent with the right optimum.
For example, using such a mutation strategy, the search indicates that the correct range for \param{INS\_POS1\_Z} should remain within $[-4.7, 0.0]$ when it searches for the lower bound. 
However, there are still incorrect configurations, specifically between $-1.0$ and $0.0$, inside this range. 
The reason is that, because of the binary search, as the first midpoint (i.e., $-2.5$) does not cause an unstable state, the search directly skips values greater than $-2.5$. 
It thus does not cover the $[-1.0, 0.0]$ space, which results in missing potentially incorrect configurations.

In the case of multiple parameters mutations, we first apply the \tool{Multi-dimensional Mutation} to determine the correct range.
After that,  based on this correct range, we leverage \system to start another search to evaluate whether there are incorrect configurations.
\system still detects $727$ potentially incorrect configurations, of which only $185$ are confirmed. 
Such result indicates that the ranges provided by the \tool{Multi-dimensional Mutation} are not correct.
In our opinion, the \tool{Multi-dimensional Mutation} is still a one-dimension mutation, because it uses binary search to mutate parameters separately but only imports the extremes of the value ranges of other parameters. 
It can be regarded as multiple one-dimensional mutations with a limited correlation between control parameters; as such, it does not consider the influence of values different from the extremes of the ranges.

\vspace{0.15cm}
\noindent
\textbf{Comparison with Respect to Correct Range Guidelines.}
The six parameters utilized in this experiment are listed in Appendix~\ref{app:a}.
We leverage \system to search incorrect configurations for these six parameters.
The search detects $1,199$ unique potentially incorrect configurations. 
Then the validation determines that $714$ out of those configurations actually lead to unstable states.
After that, we use them to generate the feasible range guidelines and choose the lowest incorrect ratio result.
Table~\ref{tab:shrink_cmp} lists the ranges obtained by three methods,
    where \tool{1} is \tool{One-dimensional mutation},
    \tool{M} is \tool{Multi-dimensional mutation},
    and \tool{GA} (our \tool{Genetic Algorithm mutation}).

\rvfuzzing method \tool{1} obtains little reduction for each parameter range; thus it is not able to rule out incorrect configurations.
\tool{M} avoids some of the incorrect configurations but still misses others.
Because our \tool{GA} greatly reduces the ranges, 
    it provides high stability. 
A special case is related to \param{INS\_POS3\_Z}; the range given by \tool{GA} is larger than the range given by \tool{M}.
But this does not mean that our range for \param{INS\_POS3\_Z} is incorrect. 
The reason is that,
    as other parameters have a smaller range, 
    \param{INS\_POS3\_Z} can have a larger range, since the validity of configurations is decided based on multiple parameters instead of a single one.
As for other parameters, 
    \param{ANGLE\_MAX} influences the flight inclination angle.
Under the influence of other parameters, 
    a too large value for \param{ANGLE\_MAX} is more likely to cause flight problems. 
A too small value for the waypoint speed \param{WPNAV\_SPEED} makes the drone more likely to freeze; 
    both \tool{M} and \tool{GA} reduce the lower part of the range for this parameter.
For \param{INS\_POS*\_Z}, the ranges returned by \tool{GA} are smaller than the ones returned by \tool{M}, and closer to default values.
In fact, changing \param{INS\_POS*\_Z} influences the position judgment of the inertial measurement unit. Therefore, this parameter should be changed carefully and should not deviate too much from the default value.
\param{PSC\_VELXY\_P} affects the output gain of the system for acceleration;
   a too large gain can easily cause drone deviations or thrust losses. 
\begin{table}[ht]
\caption{Comparison of range guideline.}
\label{tab:shrink_cmp}
\centering
\begin{threeparttable}
\begin{tabular}{c|cc|cc|cc}
        \hline
        \multirow{2}{*}{Parameter}  &   \multicolumn{2}{c|}{\tool{1}}  &  \multicolumn{2}{c|}{\tool{M}}  &  \multicolumn{2}{c}{\tool{GA}} \\
        \cline{2-7}
        ~  &  \textbf{L}  &  \textbf{U}  &  \textbf{L}  &  \textbf{U}  &  \textbf{L}  &  \textbf{U} \\
        \hline
         PSC\_VELXY\_P   &  0.1 &  6.0   &  {0.6}  &  6.0  &  1.9  &  4.2   \\
        
        INS\_POS1\_Z   &  -4.7 &  5.0   &   {-1} & 4.1  & -0.5 & 2.1  \\
        
        INS\_POS2\_Z   &  -5.0 &  5.0   &   -0.7 & 3.2  & -0.7 & 1.2  \\
        
        INS\_POS3\_Z   &  -5.0 &  5.0   &   {-0.8} & 3.0  & -0.9 & 3.2   \\
        
        WPNAV\_SPEED   &  50 &  2000   &   {300} & 2000  & 50 & 1950 \\
        
         ANGLE\_MAX   &  1000 &  8000  &   1000 & 8000  &  1100 & 4650 \\
        \hline
        \textbf{I},\textbf{V},\textbf{C} & \multicolumn{2}{c|}{699/476/1175} & \multicolumn{2}{c|}{32/18/50} & \multicolumn{2}{c}{0/7/7} \\
        \hline
\end{tabular}
\begin{tablenotes}
\footnotesize
\item[*]\tool{1}: One-dimensional mutation,
\tool{M}: Multi-dimensional mutation,
\tool{GA}: Genetic Algorithm mutation,
\textbf{L} is range lower bound,
\textbf{U} is range upper bound,
\textbf{I} is the number of incorrect configurations covered by the range guidance,
\textbf{C} is the number of correct configurations covered by the range guidance,
\textbf{V} is the number of validated configurations covered by the range guidance.
\end{tablenotes}
\end{threeparttable}
\end{table}

\vspace{0.15cm}
\noindent
\textbf{Comparison with Respect to Time Requirements.}
We analyze the time taken by \system and \tool{Multi-dimensional Mutation}.
To determine the range guideline for the six parameters,
    we started multiple simulations and took $696$ seconds to collect log data. 
The predictor takes $700$ seconds to train until reaching convergence.
The GA search takes another $156$ seconds to iterate and update its population (1,000 configurations).
Finally,
    from generating data to searching out $1,000$ incorrect configurations,
    our system totally takes $1,552$ seconds. Also,  over $85\%$ configurations are validated and detected to actually cause unstable states.  
On the contrary,
    depending on the configuration values, 
	\rvfuzzing usually takes between $20$ and $120$ seconds per-round to complete a \tool{AVC2013}.    
Even we allocate $1,552$ seconds to \tool{Multi-dimensional Mutation}, it can only carry out $78$ rounds of mission test.
In fact, the
    \tool{Multi-dimensional Mutation} search takes more than $2$ hours to estimate the guidelines for the six parameters.
In addition,
    for $20$ parameters, 
    the time taken by tool{Multi-dimensional Mutation} increases exponentially. 
Instead, the time consumption of \system is still closer to the consumption of six parameters because the time required by predictor is almost unchanged.

\subsection{Case Studies}\label{ex5}
After obtaining the fuzzing results, 
    we select several representative examples of unstable states to analyze their characteristics.

\vspace{0.15cm}
\noindent
\textbf{Deviation.}
There are two main deviation situations in our experiment: overshoot and fly away.
When flying from one waypoint to another,  the drone first accelerates to reach the target velocity and decelerates while approaching the next waypoint.
If the configuration is set improperly, the drone is unable to reduce its speed when close to the waypoint, which causes an overshoot deviation (see the video of a simulation in which the drone is not able to brake  at~\cite{dev_over}).
In comparison, fly away is much more dangerous, in that the drone keeps moving away from its mission-planned path. 
We leverage a real drone experiment to demonstrate the damage resulting from those unstable states.
The experiment sets up a simple surround flight mission and a drone configuration is set that, in the simulation test, results in a fly away deviation.
As shown in the video~\cite{dev_around}, the drone deviates from the mission-planned path.  Unfortunately,  even though we used the RC controller (remote manual controller) to manually switch its mode to \tool{land} in order to stabilize the drone and make sure it would land slowly, 
    the drone was still unable to stabilize and kept deviating. 
In fact, after checking the offline flight log,  
    when we manually switched to the \tool{land} mode, 
    the system reported a switching error indicating that it could not stabilize and land. 
In other words, if the incorrect configuration is not dealt with in time, the flight stability cannot be even corrected manually.

\vspace{0.15cm}
\noindent
\textbf{Flight Freeze.}
There are two main situations of flight freeze. 
On the one hand, an incorrect control gain parameter makes the drone fail to reach the desired position in time, or prolong the time to reach a particular waypoint or the desired position.
On the other hand, an invalid configuration affects the position and causes the drone to around-flight within a small area. 
We tested the around-flight situation with a real drone experience. 
The video at~\cite{frozen} shows that the drone keeps circling and is unable to complete the given mission.
We analyzed the offline flight log, and from the log, we could see that the drone kept flight switching between \tool{althold} (hovering fly) and \tool{land} but could not land successfully.

\vspace{0.15cm}
\noindent
\textbf{Drone Crash.}
Drone crash may be caused by a rollover during takeoff or by hitting the ground due to improper descent speed.
The video at~\cite{crash} shows an actual example of a drone crashing when taking off.

\vspace{0.15cm}
\noindent
\textbf{Potential Thrust Loss.}
This situation is mainly caused by an excessive change in angle or speed; 
    the flight controller attempts to recover its position to normal, 
    but the power of the motor cannot satisfy the requirement.
Failure in recovering from the thrust loss or stabilizing the altitude would lead to a drone crash.
In the postmortem analysis, 
    we found that changing the control parameters related to the inertial measurement unit position (e.g., \param{INS\_POS*\_Z} and the PID angle controller (e.g., \param{ATC\_ANG\_*\_P/I/D}) can force the drone into a gradually amplified oscillation.

\vspace{0.15cm}
\noindent
\textbf{Incorrect Configuration Tackling.}
Incorrect configuration tackling is the new type of error we defined. 
A control program usually contains a checking mechanism to prevent obvious errors in configurations. 
If the parameters are related to the position controller (e.g., \param{ATC\_*\_*\_P/I/D}),
    the incorrect configuration will raise a warning when the GCS tries to arm the drone.
But in fact,
    if we update the configuration after the drone take off, the problem is that the 
    the control system still accepts it,
    which ultimately drives the drone to become unstable and out of control.
The video at~\cite{tacking} shows an example in which uploading incorrect configuration during the mission causes the drone to crash.

\section{Related Work}
\label{sec:related}
\subsection{Drone Fault Detection}
A large number of drone fault detection methods/systems have been proposed to prevent errors during flights. 
Among them,
    Choi \emph{et al.}~\cite{choi2018detecting} use the control invariant to identify physical attacks against robotic vehicles.
But their approach is unable to prevent attacks that exploit the \bugs. 
To detect faulty components in a drone system, 
    G. Heredia \emph{et al.}~\cite{heredia2008sensor} construct an observer model to estimate the output in fault-free conditions from the history inputs and outputs.
Such a system utilizes the difference between actual outputs and the predicted values to detect faulty sensor and actuator components, 
    but it still does not consider the instability caused by \bugs.  
Ihab \emph{et al.}~\cite{samy2008neural} leverage the analytical redundancy between flight parameters to detect sensor faults.
Like the other approaches, it only considers external factors. 
In comparison,
    \system implements a search system to avoid incorrect configurations that are instability factors within the system.

\subsection{Learning-Based Testing}
From the perspective of learning-based testing, there are three relevant systems or methods.
NEUZZ~\cite{she2019neuzz} is a gradient-guided search strategy.
It uses a feed-forward neural network to approximate program branching behaviors and predicts the control flow edges to cover more test space,
    but the predictive model is not used for guiding testing.
ExploitMeter~\cite{yan2017exploitmeter} uses dynamic fuzzing tests with machine learning-based prediction to updates the belief in software exploitability.
Yuqi \emph{et al.}~\cite{chen2019learning} use an LSTM network to model the input-output relation of the target system and a meta-heuristic method to search for specific actuator commands that could drive the system into unsafe states.
Konstantin \emph{et al.}~\cite{bottinger2018deep} apply a deep Q-learning~\cite{mnih2013playing} algorithm to generate an optimized policy for the fuzzing-based testing of PDF processing programs.
DeepSmith~\cite{cummins2018compiler} trains a generative model with a large corpus of open source code and uses this model to produce testing inputs to examine the OpenCL compiler automatically.
These approaches make use of prior knowledge to drive the mutation input.
Similarly, 
    we introduced a machine learning model to guide the search test process. 
However, 
    our \system combines the model with the genetic algorithm to carry out a large-scale multi-parameter search.

\section{Conclusion}
\label{sec:conclusion}
Incorrect configurations of drone control parameters, set by legitimate users or worst sent by attackers, can result in flight instabilities disrupting drone missions.
In this paper,
    we propose a fuzzing-based system that efficiently and effectively detects the incorrect parameter configurations.
\system uses
    a machine learning-guided fuzzing approach that uses a predictor, a genetic search algorithm, and multi-objective optimization to detect incorrect configurations and provide correct feasible ranges.
    We have experimentally compared \system with a state-of-the art tool. The experimental results show that \system is superior to such a tool in all respects. 
    Even though our methodology has been designed for aerial drones, we believe that it can be used for other mobile devices with complex controls, such as underwater drones.


\bibliographystyle{ACM-Reference-Format}
\bibliography{sample}

\newpage
\onecolumn
\appendix
\section{Description of Parameters}
\label{app:a}

\begin{table}[ht]
\small
\caption{Parameters of control program for experiments.}
\label{tab:paramall}
\centering
\begin{tabular}{c|cccp{0.52\columnwidth}}
        \hline
        Control Module &  Parameter & Range & Default & Description.\\
        \hline
        \multirow{15}*{Controller} & PSC\_POSXY\_P & [0.50, 2.00] & 1.0 & Position controller P gain. Converts the distance (in the latitude direction) to the target location into a desired speed which is then passed to the loiter latitude rate controller.\\
       
        ~ & PSC\_VELXY\_P & [0.10, 6.00] & 2.0 & Velocity (horizontal) P gain. Converts the difference between desired velocity to a target acceleration.\\ 
       
        ~ & PSC\_POSZ\_P & [1.00, 3.00] & 1.0 & Position (vertical) controller P gain. Converts the difference between the desired altitude and actual altitude into a climb or descent rate which is passed to the throttle rate controller.\\
       
        ~ & ATC\_ANG\_RLL\_P & [0.00, 12.0] & 4.5 & Roll axis angle controller P gain. Converts the error between the desired roll angle and actual angle to a desired roll rate.\\
       
        ~ & ATC\_RAT\_RLL\_I & [0.01, 2.00] & 0.135 & Roll axis rate controller I gain. Corrects long-term difference in desired roll rate vs actual roll rate.\\
       
        ~ & ATC\_RAT\_RLL\_D & [0.00, 0.05] & 0.0036 & Roll axis rate controller D gain. Compensates for short-term change in desired roll rate vs actual roll rate.\\
       
        ~ & ATC\_RAT\_RLL\_P &  [0.01, 0.50] & 0.135 & Roll axis rate controller P gain. Converts the difference between desired roll rate and actual roll rate into a motor speed output.\\
       
        ~ & ATC\_ANG\_PIT\_P & [0.00, 12.0] & 4.5 & Pitch axis angle controller P gain. Converts the error between the desired pitch angle and actual angle to a desired pitch rate.\\
       
        ~ & ATC\_RAT\_PIT\_P & [0.01, 0.50] & 0.135 & Pitch axis rate controller P gain. Converts the difference between desired pitch rate and actual pitch rate into a motor speed output.\\
       
        ~ & ATC\_RAT\_PIT\_I & [0.01, 2.00] & 0.135 & Pitch axis rate controller I gain. Corrects long-term difference in desired pitch rate vs actual pitch rate.\\
       
        ~ & ATC\_RAT\_PIT\_D & [0.00, 0.05]& 0.0036 & Pitch axis rate controller D gain. Compensates for short-term change in desired pitch rate vs actual pitch rate.\\
       
        ~ & ATC\_ANG\_YAW\_P & [0.00, 6.00] &4.5&  Yaw axis angle controller P gain. Converts the error between the desired yaw angle and actual angle to a desired yaw rate.\\
       
        ~ & ATC\_RAT\_YAW\_P & [0.10, 2.50] & 0.18 &Yaw axis rate controller P gain. Converts the difference between desired yaw rate and actual yaw rate into a motor speed output.\\
       
        ~ & ATC\_RAT\_YAW\_I & [0.01, 1.00]& 0.018 & Yaw axis rate controller I gain. Corrects long-term difference in desired yaw rate vs actual yaw rate.\\
       
        ~ & ATC\_RAT\_YAW\_D & [0.00, 0.02] & 0 &Yaw axis rate controller D gain. Compensates for short-term change in desired yaw rate vs actual yaw rate.\\
        \hline
        \multirow{5}*{Mission} & WPNAV\_SPEED & [20, 2000] & 500 & Defines the speed in $cm/s$ which the aircraft will attempt to maintain horizontally during a Waypoint mission.\\
       
        ~ & WPNAV\_SPEED\_DN & [10, 500] & 150 & Defines the speed in $cm/s$ which the aircraft will attempt to maintain while descending during a Waypoint mission.\\
       
        ~ &  WPNAV\_SPEED\_UP & [10, 1000] & 250 & Defines the speed in $cm/s$ which the aircraft will attempt to maintain while climbing during a Waypoint mission.\\
       
         ~ & WPNAV\_ACCEL & [50, 500]& 100 &Defines the horizontal acceleration in $cm/s^2$ used during missions.\\
       
      ~ & ANGLE\_MAX & [1000, 8000] & 4500& Maximum lean angle in all flight modes.\\
        \hline
\end{tabular}
\end{table}

       
       
       
       
       
       
       
       
       
       
       
       
       
       
       
       
       
       

\begin{table}[ht]
\small
\caption{Parameters of control program for comparisons.}
\label{tab:param}
\centering
\begin{tabular}{c|cccp{0.53\columnwidth}}
        \hline
        Control Module &  Parameter & Range & Default & Description.\\
        \hline
        \multirow{1}*{Controller} &  PSC\_VELXY\_P & $[0.0, 6.0]$ & $2.0$&  Converts the difference between desired velocity to a target acceleration.\\

        \multirow{3}*{Sensor} & INS\_POS1\_Z & $[-5.0, 5.0]$& 0.0&  Z position of the first IMU accelerometer in body frame.\\
        
        ~& INS\_POS2\_Z & $[-5.0, 5.0]$&0.0& X position of the second IMU accelerometer in body frame.\\
        
        ~& INS\_POS3\_Z & $[-5.0, 5.0]$&0.0& Y position of the second IMU accelerometer in body frame.\\
        \hline
        \multirow{2}*{Mission}  & WPNAV\_SPEED & [20, 2000] & 500 & Defines the speed in $cm/s$ which the aircraft will attempt to maintain horizontally during a Waypoint mission.\\

        ~ & ANGLE\_MAX & [1000, 8000] & 4500& Maximum lean angle in all flight modes.\\
        \hline
\end{tabular}
\end{table}


        
        


\end{document}